\documentclass[final]{cvpr}

\usepackage{times}
\usepackage{epsfig}
\usepackage{graphicx}
\usepackage{amsmath}
\usepackage{amssymb}

\usepackage{multirow}
\usepackage{booktabs}
\usepackage{bbding}
\usepackage{threeparttable}
\usepackage{bm}
\usepackage[pagebackref=true,colorlinks,bookmarks=false]{hyperref}

\begin{document}

\title{HiT: Hierarchical Transformer with Momentum Contrast for Video-Text Retrieval}

\author{Song Liu\textsuperscript{1}, Haoqi Fan\textsuperscript{2}, Shengsheng Qian\textsuperscript{3,4}\thanks{Corresponding author}, Yiru Chen\textsuperscript{5}, Wenkui Ding\textsuperscript{5}, Zhongyuan Wang\textsuperscript{5} \\ 
\textsuperscript{1}Peking University, \textsuperscript{2}FAIR\\
\textsuperscript{3}National Lab of Pattern Recognition, Institute of Automation, CAS \\
\textsuperscript{4}University of Chinese Academy of Sciences\\ \textsuperscript{5}Kuaishou Technology \\
\tt\small slpku@pku.edu.cn, haoqifan@fb.com, shengsheng.qian@nlpr.ia.ac.cn \\
\tt\small \{chenyiru, dingwenkui, wangzhongyuan\}@kuaishou.com}
\maketitle

\newcommand{\hq}[1]{{\color{blue}haoqi: #1}}

\begin{abstract}
   Video-Text Retrieval has been a hot research topic with the growth of multimedia data on the internet. Transformer for video-text learning has attracted increasing attention due to its promising performance.
   However, existing cross-modal transformer approaches typically suffer from two major limitations: 1) Exploitation of the transformer architecture where different layers have different feature characteristics is limited; 2) End-to-end training mechanism limits negative sample interactions in a mini-batch.
   In this paper, we propose a novel approach named Hierarchical Transformer (HiT) for video-text retrieval. HiT performs Hierarchical Cross-modal Contrastive Matching in both feature-level and semantic-level, achieving multi-view and comprehensive retrieval results. 
    Moreover, inspired by MoCo, we propose Momentum Cross-modal Contrast for cross-modal learning to enable large-scale negative sample interactions on-the-fly, which contributes to the generation of more precise and discriminative representations.
   Experimental results on the three major Video-Text Retrieval benchmark datasets demonstrate the advantages of our method. 

\end{abstract}

\section{Introduction}
Cross-modal Retrieval \cite{tmm,ijcai20,coot,jsfusion,dual,hgr,howto100m,corrae,devise,pvse,position,frag,stacked,adversarial,sigirLiuQGZY20,aaaiQianXZFX21} has attracted the increasing attention with the aim to search the semantic similar samples from different modalities. 
Specially, the explosive growth of video contents on the internet has brought great challenges to accurate video-text retrieval. In this paper, we focus on the learning of video-text retrieval and also hope to inspire other cross-modal tasks.

\begin{figure}[t]
  \vspace{-3mm}
  \centerline{\includegraphics[width=3in]{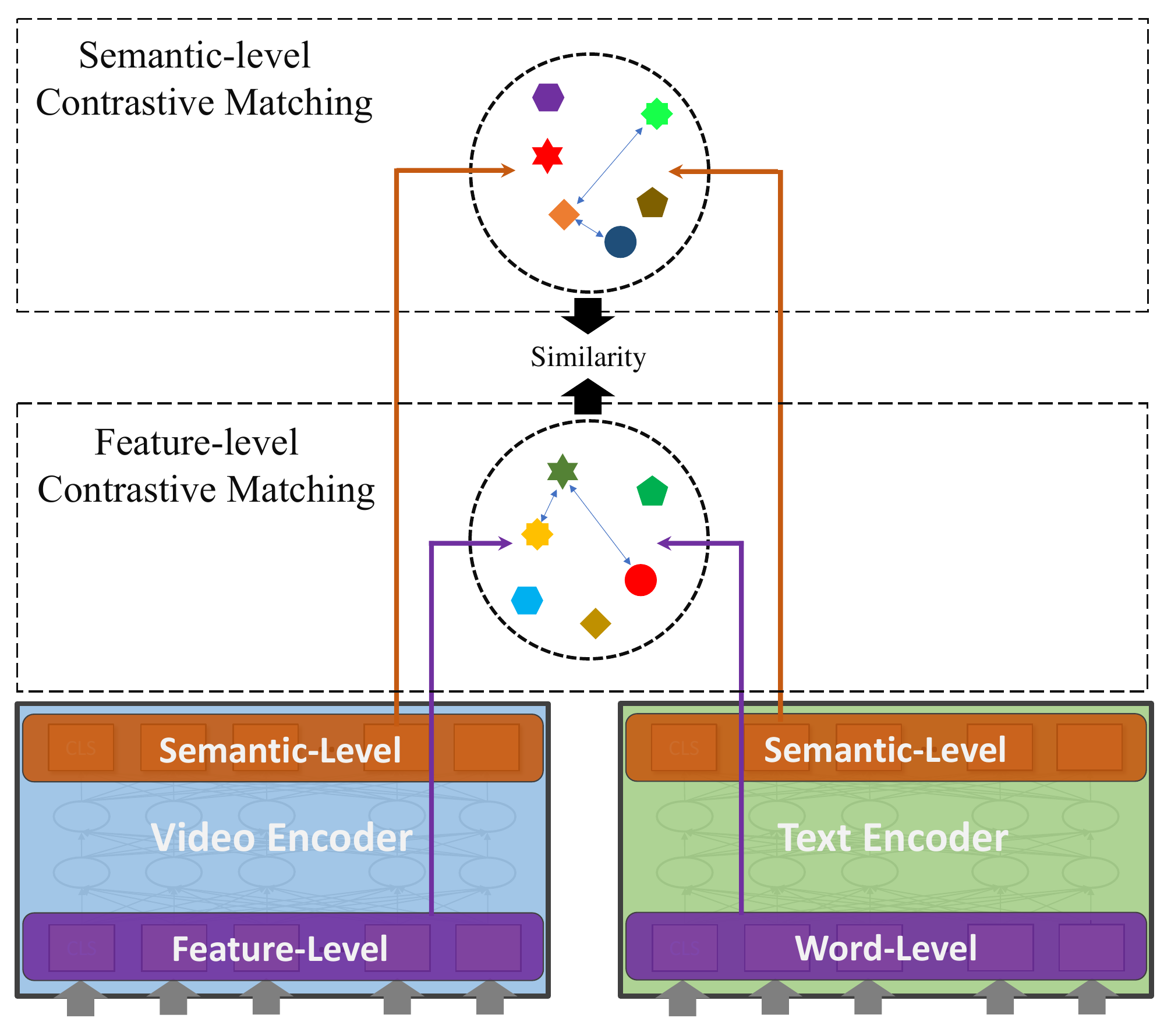}}
  \caption{Hierarchical Cross-modal Contrastive Matching consists of Feature-level and Semantic-level Contrastive Matching. \textit{Notably, Momentum Cross-modal Contrast is not shown in this figure.}}
  \label{fig1}
  \vspace{-3mm}
  \end{figure}
  
\begin{figure}[t]
  \setlength{\belowcaptionskip}{-0.2cm} 

 
   \begin{minipage}{0.25\textwidth}
    \centerline{\includegraphics[width=4cm,height=3cm]{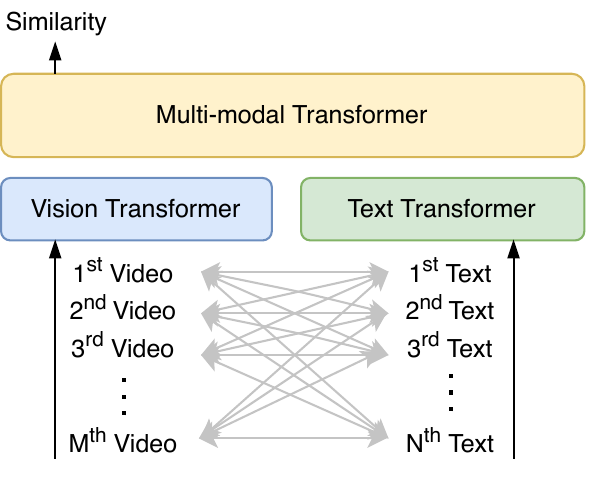}}
    \centerline{\scriptsize (a) Two-stream Architecture}
  \end{minipage}
  \hfill
  \begin{minipage}{0.2\textwidth}
    \centerline{\includegraphics[width=4cm,height=3cm]{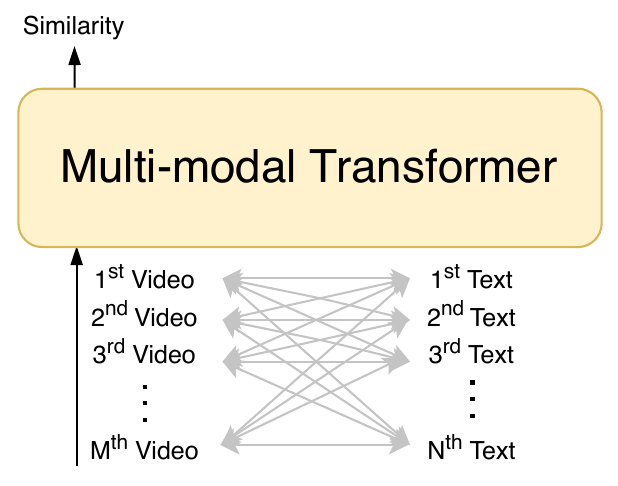}} 
    \centerline{\scriptsize (b) Single-stream Architecture}
  \end{minipage}
    \vfill
\begin{minipage}{0.5\textwidth}
    \centerline{\includegraphics[width=4.5cm,height=3.2cm]{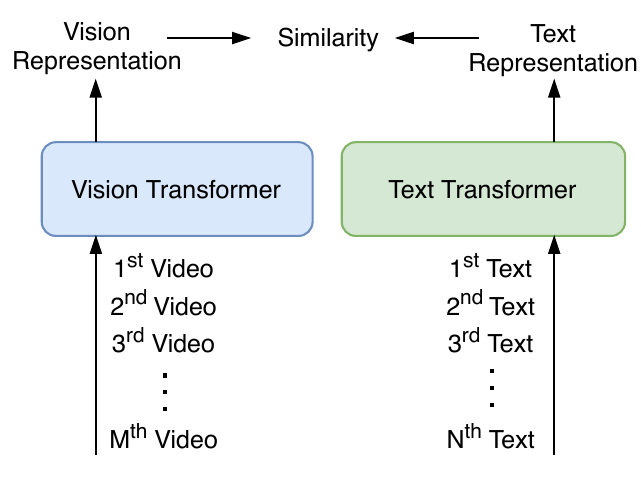}}
    \centerline{\scriptsize (c) Dual-stream Architecture}
  \end{minipage}
 
  \vbox{}
  \caption{Three types of transformer based architectures for cross-modal learning. Assuming that we have $M$ videos and $N$ texts, (a) Two-stream  and (b) Single-stream Architecture require pairwise inputs with $\mathcal{O}(MN)$ time complexity. (c) Dual-stream Architecture requires unary inputs with $\mathcal{O}(M+N)$ time complexity.  }

  \label{fig4}
  \vspace{-3mm}
  \end{figure}
 
Recent works \cite{videoBERT,actBERT,coot,iclr21,mmt} have shown that  transformer can learn high level video representations, which capture semantically meaningful and temporally long-range structures for videos. 
Notably, existing approaches for cross-modal learning can be roughly categorized as two-stream, single-stream and dual stream architectures. Two-stream architecture, as shown in Figure~\ref{fig4}-(a), utilizes a vision transformer and a text transformer to learn visual and textual representations independently, then introduces a multi-modal transformer \cite{ViLBERT,actBERT,lxmert} to achieve cross-modal information exchange. Singe-stream architecture \cite{visualBERT,unicoder,vlBERT,pixel}, as shown in Figure~\ref{fig4}-(b), fuses visual and textual representations at the initial stage of the transformer model. However, these two architectures are not suitable for large-scale cross-modal retrieval tasks, due to the requirement of pairwise inputs and $\mathcal{O}(MN)$ time complexity of intra-model information exchange. Approaches with dual-stream architecture \cite{coot,iclr21,mmt,T2VLAD} and our method, as shown in Figure~\ref{fig4}-(c), have become a recent trend for cross-modal retrieval with better efficiency, requiring a time complexity of $\mathcal{O}(M+N)$. In the line of dual-stream architecture, this paper proposes a novel transformer based method to achieve video-text retrieval, namely Hierarchical Transformer (HiT), where two contributions are jointly performed:



\textbf{Hierarchical Cross-modal Contrastive Matching.} According to the attention allocated characteristics of different layers in transformer architectures, the features in different layers focus on different views for samples \cite{visualizingbert,understanding, rediscovers,visualBERT}. For example, the  features in lower layers tend to encode more local contents with basic syntactic representations. Higher layers capture more complex semantics and usually produce higher-level semantic representations, as recent works \cite{mmt, iclr21} performed. Based on these specialities, we propose Hierarchical Cross-modal Contrastive Matching to achieve multi-view and comprehensive video-text retrieval hierarchically, which is designed as Figure~\ref{fig1}.  

\textbf{Momentum Cross-modal Contrast.} Recently, a class of  self-supervised methods for unsupervised visual representation learning \cite{momerybank,improved,he2020momentum,simclr} emphasize the necessity of large-scale negative samples.  Inspired by these works, we argue that large-scale negative sample interactions in the training process have been neglected in cross-modal contrastive learning. In this paper, we introduce MoCo \cite{he2020momentum,improved} into HiT to enable large-scale negative sample interactions on-the-fly. We name it as Momentum Cross-modal Contrast (MCC). In MCC, we build several memory banks to save a rich set of negative representations, which help broader negative sample interactions during training. However, if we utilize video and text encoders that are updated dramatically by gradient descent to generate representations for memory banks, it would result in the representation inconsistency in memory banks, thus largely affect the retrieval performance. Hence, key encoders for two modalities with momentum update (updated more smoothly) are required to maintain representation consistency.

\textbf{Contributions}:
 We propose Hierarchical Transformer (HiT) with Momentum Contrast for Video-Text Retrieval, which jointly performs Hierarchical Cross-modal Contrastive Matching and Momentum Cross-modal Contrast.
     Extensive experiments demonstrate the advantages of the proposed methods on three benchmarks, including MSR-VTT, ActivityNet and LSMDC. 
    
    


\section{Related Work}

\subsection{Video-Text Retrieval}
Video-Text Retrieval has received wide attention with the exploitation of the huge multimedia data and rich application scenarios. Several excellent works \cite{jsfusion,dual,pvse,tmm,ijcai20,coot,dual,hgr,howto100m,corrae,devise} are introduced to address this task. 
JSFusion \cite{jsfusion} proposes a joint sequence fusion model for sequential interaction of videos and texts. Dual Encoding \cite{dual} consists of mean pooling, biGRU and CNN models to encode sequential videos and texts in multiple levels. PVSE \cite{pvse} presents a polysemous instance embedding network to learn multiple and diverse representations of videos and texts for the polysemous problem. A graph-based framework is proposed in \cite{graph} for matching between movie segments and synopsis paragraphs, which takes into account both the flow of events and the interactions among characters. HGR \cite{hgr} is a Hierarchical Graph Reasoning model, which decomposes video-text matching into global-to-local levels and disentangles texts into a hierarchical semantic graph with three levels of events, actions and entities. 



\begin{figure*}[t]
  \centerline{\includegraphics[width=7in]{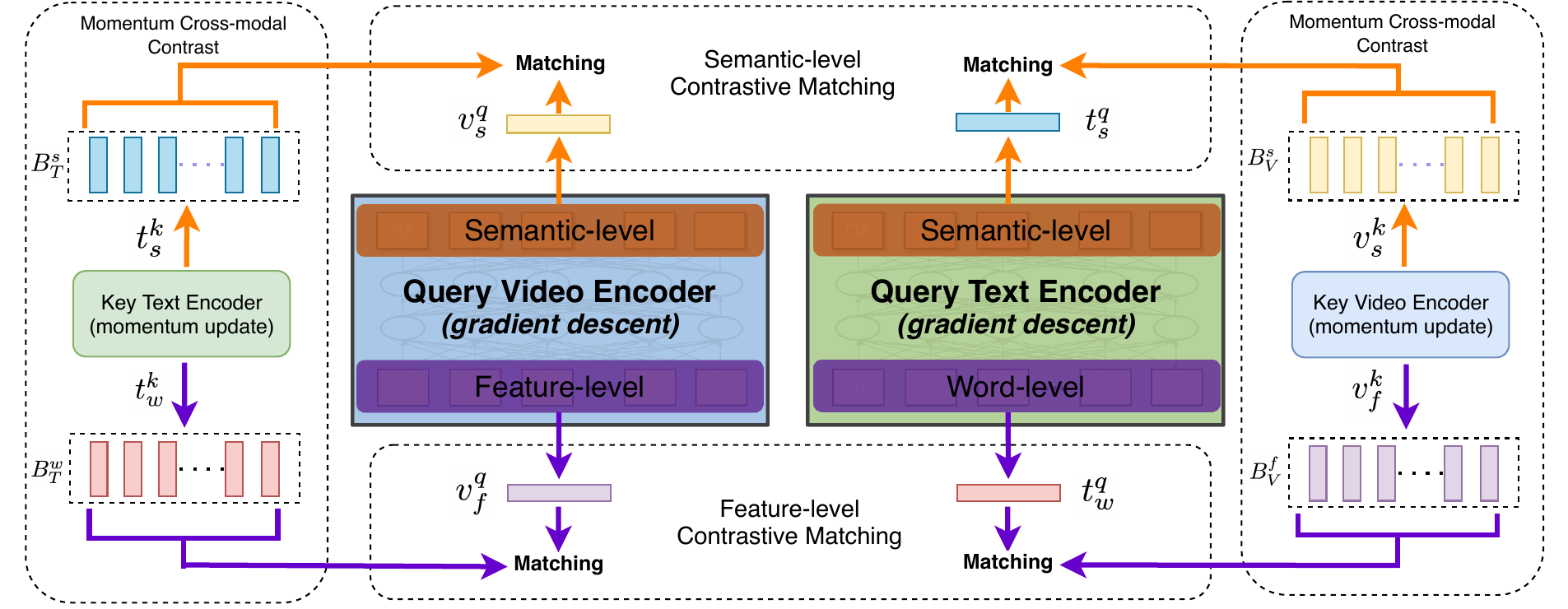}}
  \caption{The pipeline of our method. All encoders adopt transformer based architectures. \textbf{Video Encoding}: Query Video Encoder and Key Video Encoder. \textbf{Text Encoding}: Query Text Encoder and Key Text Encoder. \textbf{Momentum Cross-modal Contrast}: Four memory banks are built to save the key representations from two level of two modalities. Two query encoders are  updated by gradient descent and two key encoders are momentum updated. \textbf{Hierarchical Cross-modal Contrastive Matching}: Semantic-level Contrastive Matching is performed between query video (text) semantic-level representations and key text (video) semantic-level representations in memory banks. Feature-level Contrastive Matching is performed between query video feature-level (text word-level) representations and key text word-level (video feature-level ) representations in memory banks.}
  
  \label{fig3}
  
  \end{figure*}

\subsection{Video-Text Learning with Transformer}
BERT \cite{BERT} is a transformer based representation model for natural language process tasks. It evolves a line of works that learn a universal language encoder by pre-training with language modeling objectives. 
Recently, several attempts \cite{visualBERT,ViLBERT,vlBERT,lxmert,mmt,videoBERT,actBERT,unicoder,pixel, CLIPBERT,confsigirQianWHFX21} have been made which utilize BERTs and transformers as the backbones for cross-modal tasks.
In video-text learning tasks, VideoBERT \cite{videoBERT} transforms a video into spoken words paired with a series of images and applies a transformer to learn joint representations. ActBERT \cite{actBERT} learns a joint video-text representation that uncovers global and local visual clues from paired video sequences and text descriptions. Both the global and the local visual signals interact with the semantic stream mutually. MMT \cite{mmt} proposes the  multi-modal transformer which processes features extracted from different modalities at different moments in videos, such as video, audio and speech. COOT \cite{coot} proposes a hierarchical model that exploits long-range temporal context producing the video/text embeddings based on hierarchically interactions between local and global contexts. Support-set \cite{iclr21} incorporates a auxiliary generative task, \textit{i.e.,} cross-captioning task, to alleviate mismatching problems existed in recent works. Very recently, T2VLAD \cite{T2VLAD} uses a paradigm of global-local alignment to perform video retrieval. It obtains the global similarities by calculating the similarities multiple times between video-related and text features. For obtaining local similarities, they need to cluster the local features into several shared centers firstly, and calculate the similarities between local features and cluster centers. Though it also performs hierarchical matching as HiT, it performs their idea in a more complicated way.


\subsection{Contrastive Learning}
Contrastive Learning \cite{simclr,he2020momentum,improved,CMC,SupCon,infomin,PIRL,unsupervised,simsiam,BYOL} has made the remarkable progress in unsupervised visual representation learning. We introduce several representative contrastive learning mechanisms that benefit from the optimization with negative samples. \textit{End-to-end} mechanism uses samples in the current mini-batch, where one can use its augmented views  as positive samples and consider other samples in the current batch as negatives. \textit{Memory bank} \cite{momerybank} mechanism uses the representations sampled from a memory bank to conduct broader negative sample learning. However, the representations in the memory bank are from very different encoders all over the past epoch and they are less consistent. MoCo \cite{improved,he2020momentum} improves the memory bank mechanism by using a momentum-updated key encoder to generate the large-scale negative representations for the memory bank which can maintain better representations’ consistency.
SimCLR \cite{simclr} shows that contrastive learning in unsupervised visual representation learning benefits from large batch size negatives, stronger data augmentation and introducing the learnable nonlinear transformation, \textit{i.e.,} using projection heads. Though recent works \cite{simsiam, BYOL} show that contrastive learning can achieve decent performance even without negatives by using a momentum encoder \cite{BYOL} or stop gradient operation \cite{simsiam} to prevent collapse solutions, our HiT in video-text retrieval and \cite{he2020momentum,simclr, improved,momerybank, mochi} in visual representation learning indeed benefit from the large-scale negative sample learning. The effects of cross-modal learning without negatives are not involved in this paper.

\section{Problem Definition}

For the video-text retrieval task, we are given $M$ videos $V=\{V_i\}_{i=0}^{M-1}$ and $N$ captions $T=\{T_i\}_{i=0}^{N-1}$. 
Each video has several kinds of expert embeddings to represent videos in multiple views, e.g., motion, appearance and audio. Each caption is represented by the natural language in English. Formally, the target of our methods for video-text retrieval is to obtain two \textit{query encoders} $f$: $V \rightarrow \textbf{Z}=\{Z_i\}_{i=1}^L$ and $g$: $T \rightarrow \textbf{Z}=\{Z_i\}_{i=1}^L$ jointly, where $f$ and $g$ are for video and text domains respectively, and $\textbf{Z}$ consists of $L$ common embedding spaces. In the common embedding spaces, cross-modal samples are represented by a series of compact embeddings.  Meanwhile, the distance among similar cross-modal samples are smaller than that of among dissimilar cross-modal samples in the common embedding spaces. The constraint can be formulated as follows:
\begin{equation}
\begin{aligned}
d(f(V_i), g(T_i)) \leq d(f(V_i), g(T_j)) \; \textit{s.t.} \; i \neq j
\end{aligned}
\end{equation}
where $d(\cdot, \cdot)$ is the distance measurement. The overall similarity between two cross-modal samples is decided by hierarchical contrastive matching results.

\section{Hierarchical Transformer}

Figure~\ref{fig3} illustrates the structure of the Hierarchical Transformer (HiT) for video-text retrieval.  
For video encoding, there are \textit{Query Video Encoder} and \textit{Key Video Encoder}. Both two video encoders utilize the same architecture. For text encoding, there are \textit{Query Text Encoder} and \textit{Key Text Encoder} that adopt the same architecture. Notably, Siamese encoders, \textit{a.k.a.,} key encoders, are shown for the utilization of Momentum Cross-modal Contrast (MCC), which will be discussed later. There are only two query encoders left if we remove MCC, as shown in Figure~\ref{fig1}. 

\begin{figure}[h]
  \centerline{\includegraphics[width=3in]{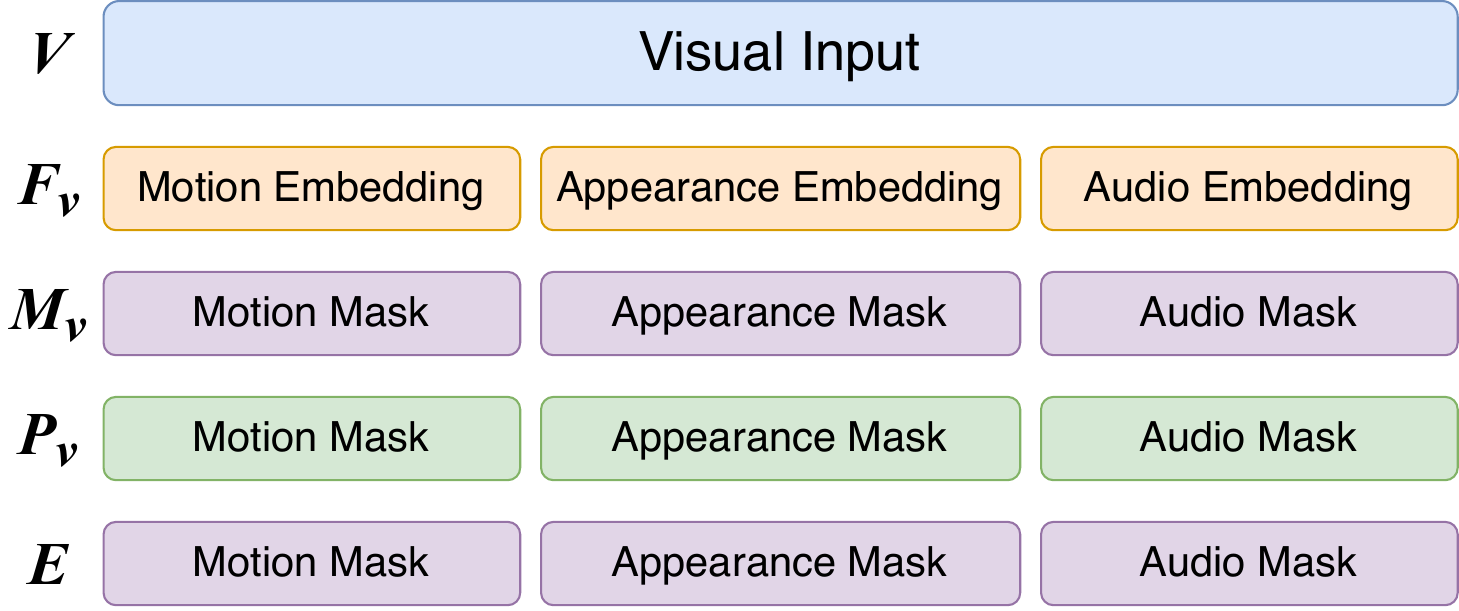}}
  \caption{The visual input of video encoders.}
  \label{fig2}
  \vspace{-3mm}
  \end{figure}

\subsection{Video Encoders} \label{video_encoders}

The video encoders, including query and key video encoders, are designed as transformer based architectures. We transform the raw visual features into a discrete sequence of tokens as inputs. To this end, we generate a sequence of pre-trained video-related features, including motion, appearance and audio features,  to obtain \textit{Visual Embeddings} $\textbf{F}_v$ as the inputs. \textit{Visual Segment Masks} $\textbf{M}_v$ and \textit{Visual Position Embeddings} $\textbf{P}_v$ are needed to indicate the real numbers and positions of input features respectively. We append \textit{Expert Embeddings} $\textbf{E}$ to identify the attending expert. The final visual input $\textbf{V}$ can be formulated as follows, also shown in Figure~\ref{fig2}:
\begin{equation}
    \textbf{V}= \textbf{F}_v + \textbf{M}_v + \textbf{P}_v + \textbf{E}
\end{equation}

\noindent
\textbf{$\bullet$ Video Feature-level Feature.} As studied in \cite{view1, view2,view3}, in the transformer based architectures, the features in lower layers capture low-level patterns that describe basic syntactic information. We obtain these visual token features in the first layer of the query video encoder and the key video encoder. Then we do \textit{Average Pooling} and \textit{Nonlinear Projection} for them and obtain $v_f^{q} \in \mathbb{R}^{D_v}$ and $v_f^{k} \in \mathbb{R}^{D_v}$, respectively. MLPs are adopted as the nonlinear projection heads to do nonlinear transformations. \cite{simclr} has proved that a nonlinear projection head can improve the representation quality of the layer before it. 

\noindent
\textbf{$\bullet$ Video Semantic-level Feature.} Higher layer features in transformer based architectures capture higher-level representations with  more complex semantic meanings. We do average pooling for the contextual tokens in the last layer to represent the semantic-level features. Then two projection heads are used to do nonlinear transformations for obtaining $v_s^{q}\in \mathbb{R}^{ D_v}$ and $v_s^{k}\in \mathbb{R}^{ D_v}$ generated by the query video encoder and the key video encoder respectively.

\subsection{Text Encoders} \label{text_encoders}

We leverage \texttt{BERT-base-uncased} \cite{bertbase} as the text encoders and fine-tune it.  It's worth noting that the video features are generated by pre-trained deep neural networks and already have higher level semantic representation ability. While the text modality has different inherent complexity from the video modality and needs more transformer blocks to model semantic relations among words. Thus, text encoders are deeper than video encoders.

Each word in a caption will be embedded into a word embedding vector and we obtain \textit{Token Embeddings} $\textbf{F}_t$. \texttt{[CLS]} and \texttt{[END]} are embedded into the first and last positions. \textit{Text Segment Mask} $\textbf{M}_t$ is needed to indicate the real length of the input sequence. 
\textit{Text Position Embedding} $\textbf{P}_t$ is used to represent the word indexes of the input sequence in text encoders. The final input for text encoders is defined as:
\begin{equation}
    \textbf{T}= \textbf{F}_t + \textbf{M}_t + \textbf{P}_t
\end{equation}


\noindent
\textbf{$\bullet$ Text Word-level Feature.} We obtain text word-level features from the first layer of query text encoder and key text encoder. Similar to the acquisition of video feature-level features, we perform average pooling and utilize two projection heads to do nonlinear transformations and obtain $t_w^{q}\in \mathbb{R}^{D_t}$, $t_w^{k}\in \mathbb{R}^{D_t}$ .

\noindent
\textbf{$\bullet$ Text Semantic-level Feature.} The average pooling of token features from the last layer are referred as text semantic-level features. These contextual tokens represent the higher-level meaning of the given caption. Two projection heads are used to do nonlinear transformations for obtaining $t_s^{q} \in \mathbb{R}^{ D_t}$ and $t_s^{k} \in \mathbb{R}^{ D_t}$.

\subsection{Momentum Cross-modal Contrast}

The end-to-end training mechanism as most methods implemented largely limits the negative sample interactions. To enable large-scale negative sample interactions for generating more precise and discriminative representations, Momentum Cross-modal Contrast (MCC) is proposed. Four memory banks are firstly built as queues for saving negative representations dynamically. 

\noindent
\textbf{$\bullet$ Text Memory Banks.} Text memory banks, including $B_T^{w}$ for saving key text word-level features and $B_T^{s}$ for saving key text semantic-level features, are built as two queues. In each training iteration, the current mini-batch key text representations $t_w^{k}$ and $t_s^{k}$ encoded by the key text encoder will be enqueued into $B_T^{w}$ and $B_T^{s}$, and the oldest mini-batch will be dequeued.
The key text representations in $B_T^{w}$ and $B_T^{s}$ will be used to calculate the loss with the current mini-batch video representation $v_{f}^q$ and $v_{s}^{q}$  encoded by the query video encoder. 

\noindent
\textbf{$\bullet$Video Memory Banks.} Similarly, video memory banks $B_V^{f}$ for saving key video feature-level features $v_f^{k}$, and $B_V^{s}$ for saving key video semantic-level features $v_s^{k}$ are built. 

\textit{Moreover, to maintain the representation consistency in the memory banks, two key encoders, which perform momentum update \cite{he2020momentum, improved}, are required.} We denote $\theta_q^v$ and $\theta_k^v$ as the parameters of the query and key video encoders. $\theta_q^t$ and $\theta_k^t$ are the parameters of the query and key text encoders. We formulate the momentum update for $\theta_k^v$ and $\theta_k^t$  as:
\begin{equation}
\begin{aligned}
    \theta_k^v \leftarrow m \theta_k^v + (1-m)\theta_q^v \\
    \theta_k^t \leftarrow m \theta_k^t + (1-m)\theta_q^t \\
\end{aligned}
\end{equation}
where $m \in [0, 1)$ is a momentum coefficient, which is a relatively large value. We set $m=0.999$ in this paper. The parameters $\theta_q^v$ and $\theta_q^t$ are updated by back-propagation. The momentum update makes $\theta_k^v$ and $\theta_k^t$ evolve more smoothly than $\theta_q^v$ and $\theta_q^t$. As a result, though the key representations in the memory banks are encoded by different encoders (in different mini-batches), the difference among these encoders will be small.

\subsection{Hierarchical Cross-modal Contrastive Matching}


We propose hierarchical cross-modal contrastive matching for video-text retrieval learning. Specifically, 
we utilize video feature-level features and text word-level features for feature-level contrastive matching. The video and text semantic-level features are used for semantic-level contrastive matching.

\textbf{Feature-level Contrastive Matching.} For the view of retrieving texts with videos, we get \textit{positive similarity} $s^{vt+}$ by calculating cosine similarity between $v_f^q$ and $t_w^k$. Then, we obtain \textit{negative similarity} $S_{vt-}=\{s_{i}^{vt-}\}_{i=1}^{K_t}$ by calculating cosine similarity among $v_f^q$ and all key text representations in $B_T^w$. Thus, we achieve $S_{vt} = \{s^{vt+}\}\cup S_{vt-} = \{ s^{vt}_{i} \}_{i=1}^{1+K_t}$, where $K_t$ is the queue size of $B_T^w$. Similarly, for the view of retrieving videos with texts, we get $S_{tv} = \{ s^{tv+} \}\cup S_{tv-} = \{ s^{tv}_{i} \}_{i=1}^{1+K_v}$, where $K_v$ is the queue size of $B_V^f$.
The InfoNCE \cite{infonce}, a form of contrastive loss functions, is adopted as our objective function for feature-level contrastive matching:
\begin{equation}
\begin{aligned}
    \mathcal{L}_{1} = - log \dfrac{\exp(s^{vt+} / \gamma)} { \sum_{i=1}^{1+ K_t} \exp(s_{i}^{vt} / \gamma )} 
    - log \dfrac{\exp(s^{tv+} / \gamma)} { \sum_{i=1}^{^{1+K_v}} \exp(s^{tv}_i/ \gamma )}
 \label{b}
\end{aligned}
\end{equation}
where $\gamma$ is a temperature hyper-parameter, which is set to 0.07 in this paper.

\textbf{Semantic-level Contrastive Matching.} Similarly, we achieve positive and negative similarity $C_{vt}= \{ c^{vt+} \}\cup C_{vt-} = \{ c^{vt}_{i} \}_{i=1}^{1+K_t}$ and $C_{tv}=\{c^{tv+} \}\cup C_{tv-} = \{ c^{tv}_{i} \}_{i=1}^{1+K_v}$. The objective function of semantic-level contrastive matching is defined as:

\begin{equation}
\begin{aligned}
    \mathcal{L}_{2} = - log \dfrac{\exp(c_{vt+} / \gamma)} {\sum_{i=1}^{1+K_t} \exp(c^{vt}_{i} / \gamma )} - log \dfrac{\exp(c_{tv+} / \gamma)} {\sum_{i=1}^{^{1+K_v}} \exp(c^{tv}_{i} / \gamma )}
 \label{d}
\end{aligned}
\end{equation}

Thus, the overall objective function is $\mathcal{L}$:
\begin{equation}
\mathcal{L} = \alpha \mathcal{L}_1 + \beta \mathcal{L}_2 
\label{L}
\end{equation}
where $\alpha$ and $\beta$ are two hyper-parameters to balance two objectives. We set both $\alpha$, $\beta$ to 1 in our experiments.

\begin{table*}[t]

\setlength{\belowcaptionskip}{-0.05cm} 
  \centering

  \caption{The experimental results on MSR-VTT. Larger R@1,R@5,R@10 and smaller MedR indicate better retrieval performance.}

  \begin{tabular}{c|cccc|cccc|c}
  \toprule[1.5 pt] 
  
  \multirow{2}{*}{\textbf{Methods}} & \multicolumn{4}{c|}{\textbf{Video-to-Text Retrieval}}  &\multicolumn{4}{|c|}{\textbf{Text-to-Video Retrieval}} &
  \multirow{2}{*}{\textbf{rsum}}
   \\ \cline{2-5}\cline{6-9}
    &{R@1} & {R@5} & {R@10} & {MedR} &{R@1} & {R@5} & {R@10} & {MedR}  \\

  \hline
  
 
 {AM \cite{AM}} &{6.8} &{18.1} &{26.5} &{42}&{7.0} &{18.1} &{27.0}& {40} & {103.5} \\
 
{LJE \cite{LJE}} & {9.2}& {27.6} &{39.1} &{22} &{6.9}& {22.5} &{29.8} &{32} &{134.9}  \\

{ActBERT \cite{actBERT}} & - & - & - & - & 8.6 & 23.4  & 33.1 & 36 & - \\

  {JSFusion \cite{jsfusion}} & {9.5} & {28.6} &{40.2} & {18}  & {9.6} &{29.8} & {42.1} & {20} & {159.8}  \\
  
  {HowTo100M \cite{howto100m}} & {12.2} & {33.5} &{47.5} & {13} & {12.6} &{36.2} & {48.1} & {13} &{190.2} \\ 
  {CE \cite{CE}} & {20.9} & {48.8} &{62.4} & {6} & {20.6} & { 50.3} & {64.0} & {5.3} & {267.0}\\
  
   {MMT  \cite{mmt}}  &{24.4} & {56.0} & {67.8}&{4} & {24.6}  &{54.0} & {67.1}& {4} &{293.9}  \\
   
  {SUPPORT-SET \cite{iclr21}}  & {26.6} &{55.1} & {67.5} & {3}& {27.4}  &{56.3} & {67.7}& {3} & {300.6}  \\
 
  \hline

  {\textbf{HiT}} & \textbf{28.8} & \textbf{60.3} & \textbf{72.3} &\textbf{3}& \textbf{27.7} & \textbf{59.2} &\textbf{72.0}  & \textbf{3}  & \textbf{320.3}\\
  
  \hline
  \hline
  
  {HowTo100M \cite{howto100m}}  & {16.8} & {41.7} &{55.10} &{8}& {14.9}  &{40.2} & {52.8}& {9}  & {221.5} \\
  
  {NoiseEstimation \cite{Noise}}  &{-}  &{-} & {-}& {-} &  {17.4} & {41.6} &{53.6} &{8} & {-} \\

  {UniVL \cite{UniViLM}} &{-}  &{-} & {-}& {-}  &  {21.2} & {49.6} &{63.1} &{6} & {-} \\
  {AVLnet \cite{AVLnet}}   & {28.5} & {54.6} &{65.2} &{4} & {27.1}  &{55.6} & {66.6}& {4} & {297.5} \\

{MMT \cite{mmt}}  & {27.0} & {57.5} &{69.7} &{3.7}& {26.6}  &{57.1} & {69.6}& {4} & {307.5}  \\

  {SUPPORT-SET \cite{iclr21}} & {28.5} & {58.6} & {71.6} &{3}& {30.1} & {58.5} &{69.3}  & {3}  & {316.6} \\
  \hline
  {\textbf{HiT Pre-trained on HT100M}} & \textbf{32.1} & \textbf{62.7} & \textbf{74.1} &\textbf{3}& \textbf{30.7} & \textbf{60.9} &\textbf{73.2}  & \textbf{2.6}  & \textbf{333.7}\\

  \toprule[1.5 pt]
  
  \end{tabular}
   
 \label{tab1}
 
 \end{table*}
 
 \begin{table}[t]

\setlength{\belowcaptionskip}{-0.05cm} 
  \centering

  \caption{Text-to-video retrieval results on ActivityNet.}

  \begin{tabular}{c|cccc}
  \toprule[1.5 pt]

   {\textbf{Methods}} &{R@1} & {R@5} & {R@50} & {MedR}  \\

  \hline
  

  {FSE \cite{HSE}}  & 18.2 & 44.8 & 89.1 & 7.0 \\
  
   {CE \cite{CE}} & 18.2 & 47.7 & 91.4 &  6.0   \\
 
 {HSE \cite{HSE}}  & 20.5 & 49.3 & - & -   \\
 
 {MMT \cite{mmt}} & 22.7 & 54.2 & 93.2 & 5.0  \\
 
 {SUPPORT-SET \cite{iclr21}} & 26.8 & 58.1 & 93.5 & \textbf{3.0} \\
  \hline
{\textbf{HiT}} & \textbf{27.7} & \textbf{58.6} &\textbf{94.7}  & {4.0}\\
 
  {\textbf{HiT Pre-trained}} &\textbf{29.6} & \textbf{60.7} &\textbf{95.6}  & \textbf{3.0} \\

  \toprule[1.5 pt]
  
  \end{tabular}
   
 \label{tab2}
 
 \end{table}

\begin{table}[t]

\setlength{\belowcaptionskip}{-0.05cm} 
  \centering

  \caption{Text-to-video retrieval results on LSMDC.}

  \begin{tabular}{c|cccc}
  \toprule[1.5 pt]

   {\textbf{Methods}} &{R@1} & {R@5} & {R@10} & {MedR}  \\

  \hline
  

  {CT-SAN \cite{CT-SAN}} &  5.1 & 16.3 & 25.2 & 46 \\
 
 {JSFusion \cite{jsfusion}}  & 9.1 & 21.2 & 34.1 & 36    \\
 
 {CCA \cite{CCA}} & 7.5 & 21.7 & 31.0 & 33  \\
 
 {MEE \cite{MEE}} & 9.3 & 25.1 & 33.4 & 27    \\
 
  {MEE-COCO \cite{MEE}} & 10.1 &25.6 &34.6 &27  \\
  {CE \cite{CE}} & 11.2 & 26.9 & 34.8 & 25.3 \\
   {MMT \cite{mmt}} & 13.2 &29.2 &38.8 &21.0 \\

  \hline
{\textbf{HiT}} & \textbf{14.0} & \textbf{31.2} &\textbf{41.6}  & \textbf{18.5}\\


  \toprule[1.5 pt]
  
  \end{tabular}
  
 \label{tab3}
 \vspace{-3mm}
 \end{table}

\section{Experiments}

\subsection{Datasets and Evaluation Metrics}
We adopt video-text retrieval experiments on three datasets. Pre-training experiments are conducted on HowTo100M \cite{howto100m}.

\noindent
\textbf{$\bullet$ MSR-VTT} \cite{msrvtt} contains 10,000 videos, where each video is annotated with 20 captions in English. We follow the training protocol defined in \cite{mmt, CE,howto100m} to evaluate on text-to-video and video-to-text retrieval tasks on the 1k-A testing split with 1,000 video or text candidates defined by \cite{jsfusion}.

\noindent
\textbf{$\bullet$ ActivityNet Captions} \cite{acitivity} consists of 20K YouTube videos temporally annotated with sentence descriptions. We follow the approach of \cite{sun2019learning,mmt}, where all the descriptions of a video are concatenated to form a paragraph. The training set has 10,009 videos. We evaluate our video-paragraph retrieval on the “val1” split (4,917 videos). 

\noindent
\textbf{$\bullet$ LSMDC} \cite{LSMDC} contains 118,081 short video clips ($\sim$45s) extracted from 202 movies. Each clip is annotated with a caption, extracted from either the movie script or the audio description. The testing set is composed of 1,000 videos, from movies not present in the training set.

\noindent
\textbf{$\bullet$ Metric.} We measure the retrieval performance with common metrics in information retrieval, including Recall at K (R@K and K=1, 5, 10), and Median Rank (MedR). R@K is the percentage of test queries that at least one relevant item is found among the top-K retrieved results. The MedR measures the median rank of correct items in the retrieved ranking list, where lower score indicates a better model. We also take the sum of all R@K as rsum to reflect the overall retrieval performance.

\begin{table*}[t]
\setlength{\belowcaptionskip}{-0.05cm}

  \centering
  \caption{Ablation study on MSR-VTT to investigate the contributions of Momentum Cross-modal Contrast.}
  \begin{tabular}{c|ccc|ccc|ccc|c}
  \toprule[1.5 pt] 
  
  \multirow{2}{*}{\textbf{Methods}} &
  \multicolumn{3}{c|}{\textbf{Memory Bank}} & \multicolumn{3}{c|}{\textbf{Video-to-Text Retrieval}} &\multicolumn{3}{c|}{\textbf{Text-to-Video Retrieval}} &
  \multirow{2}{*}{\textbf{rsum}}
   \\ \cline{2-10}
    & {Use} &{Qk} &{Qv} &{R@1} & {R@5} & {R@10} &{R@1} & {R@5} & {R@10}  
    \\ 
  
  \hline
  
  {HiT \textit{w/o MCC}} & {\XSolidBrush} & {-} &{-} & {27.1}& {55.3 } & {68.3} & {27.0 } &{58.0} & {70.8} & {306.5}  \\
  
   {HiT \textit{w MCC}} & {\CheckmarkBold} & {256} &{256} & 26.9	&56.1 & 69.0 &	27.0 &	58.6 &	71.0  &{308.6} \\
  
  {HiT \textit{w MCC}} & {\CheckmarkBold} & {512} &{512}& 27.6 &	58.3 &	{70.0}&	27.4 &	58.7 &	70.8  &{312.8} \\
  
  {HiT \textit{w MCC}} & {\CheckmarkBold} & {1,024} &{1,024}& {27.7}&{57.9}	&{70.3} &{27.3}&	\textbf{59.7}&	{71.8} &{314.7}\\

  {HiT \textit{w MCC}} & {\CheckmarkBold} & {2,048} &{2,048}&28.0&	59.6&{71.9} &	27.4  &	59.0 &	71.5 &{317.4}\\
  
  {HiT \textit{w MCC}} & {\CheckmarkBold} & {4,096} &{4,096}& \textbf{28.8} & \textbf{60.3} & {72.3} & \textbf{27.7} & {59.2} & \textbf{72.0}  & \textbf{320.3} \\
  
  {HiT \textit{w MCC}} & {\CheckmarkBold} & {8,192} &{8,192}& 28.1 &	58.9 &	\textbf{72.5} &	27.0 &	58.7 &	71.0 &{316.2}\\
  
  \toprule[1.5 pt]
   
\end{tabular}
 \label{tab5}
 \vspace{-3mm}
 \end{table*}


\subsection{Implementation Details}

\noindent
\textbf{$\bullet$ Pre-trained Features.} We follow MMT \cite{mmt} to conduct pre-trained feature extraction. Motion features are extracted from S3D \cite{s3d} trained on the Kinetics action recognition dataset. Audio features are extracted from VGGish model \cite{audio} trained on YT8M. Appearance features are extracted from the final global average pooling layer of SENet-154 \cite{senet} trained on ImageNet. 

For MSRVTT and LSMDC, we use all motion, appearance and audio experts. We employ 30 features for each type of visual features as the visual input, and the 25 first words from captions as the text input. For HowTo100M and ActivityNet, we use motion and audio experts, each of which has 100 features as the visual input, and the first 100 words as the text input.

\noindent
\textbf{$\bullet$ Backbone.} For text encoders, we use 12-layer \texttt{BERT-base-uncased} \cite{bertbase} and fine-tune it. Video encoders have 4 transformer layers with 4 attention heads. The hidden size and the intermediate size are set to 512 and 3,072, respectively. 
We set the hidden size of projection heads to 8,192. $D_v$ and $D_t$ are both set to 2,048. The ReLU is used as the activation function and BN layers are used in hidden layers. 

\noindent
\textbf{$\bullet$ Optimization.} The initial learning rate is set to 2e-5 and the network is optimized by AdamW \cite{adamw} optimizer. 
The 10\% proportion of warm up and cosine decay are used for scheduling the learning rate. 
The batch size is 128 and we train 40 epochs. 
All experiments are conducted on NVIDIA 3090Ti GPUs. 

\noindent
\textbf{$\bullet$ $K_v$ and $K_t$ in MCC .} For MSR-VTT, we report retrieval results when we set $K_v$ and $K_t$ to 4,096. $K_v$ and $K_t$ in ActivityNet are set to 512. In LSMDC, $K_v$ and $K_t$ are 1,024. We set $K_v$ and $K_t$ to 8,192 in HowTo100M. These numbers should vary with the batch size.



  
    
 

\subsection{Compare to state of the art}

The Table~\ref{tab1}-\ref{tab3} present the retrieval results of HiT on MSR-VTT, ActivityNet Captions and LSMDC. We also compare HiT with other state-of-the-art methods. 

As shown in the results, HiT outperforms all comparison methods by a clear margin. For MSR-VTT, we report video-to-text retrieval and text-to-video retrieval results. In particular, our retrieval performance at rsum is 320.3, exceeding recent state-of-the-art methods \cite{iclr21} by a margin of 19.7. It well reflects the overall retrieval quality of HiT. With pre-training on HowTo100M, HiT further boosts the retrieval performance. For ActivityNet Captions and LSMDC, we report the retrieval performance in terms of text-to-video retrieval. HiT still outperforms comparison methods. We find that the growth of retrieval performance benefits from the proposed components, including Hierarchical Cross-modal Contrastive Matching and Momentum Cross-modal Contrast. To demonstrate the effectiveness and robustness of two components, we exhaustively and comprehensively ablate our method in the following sections. 

\section{Ablation Study}

\textbf{Hierarchical Cross-modal Matching.}
As mentioned above, we use token features from the first layers to perform Feature-level Contrastive Matching while token features from the last layers are adopted for Semantic-level Contrastive Matching. In this section, we design several variants to verify the impacts of Hierarchical Cross-modal Contrastive Matching. Note that we do not perform MCC for efficiency in this .

\noindent
\textbf{$\bullet$ HiT}-\textit{sl}. We only implement semantic-level matching while feature-level matching is removed.

\noindent
\textbf{$\bullet$ HiT}-\textit{fl}. Only feature-level matching is implemented.

\noindent
\textbf{$\bullet$ HiT}-\textit{4-level}. \textit{To investigate the potential of hierarchical matching for transformer architectures, contrastive matching with respect to more levels is conducted.} Since a text encoder has 12 transformer blocks and a video encoder has 4 blocks, except feature-level (between \textit{layer-1 in text encoder} and \textit{layer-1 in video encoder}) and semantic-level (between \textit{layer-12 in text encoder} and \textit{layer-4 in video encoder}), we append contrastive matching with more levels between \textit{layer-5 in text encoder} and \textit{layer-2 in video encoder}, \textit{layer-9 in text encoder} and \textit{layer-3 in video encoder}. 

\noindent
\textbf{$\bullet$ HiT}-\textit{3-level-a}. We append contrastive matching  between \textit{layer-9 in text encoder} and \textit{layer-3 in video encoder}.

\noindent
\textbf{$\bullet$ HiT}-\textit{3-level-b}. Contrastive matching between \textit{layer-5 in text encoder} and \textit{layer-2 in video encoder} is appended. 
    
\noindent
\textbf{$\bullet$ HiT}. Original HiT in Table~\ref{tab1}.

Table~\ref{abation1} presents the ablation results on MSR-VTT in text-to-video retrieval. We find that using more levels to conduct contrastive matching is able to obtain clear improvements. However, n-level matching requires n times retrieval during inference. In addition, significant improvements are not shown in 3-level and 4-level matching results. For the sake of retrieval efficiency and efficient training with Momentum Cross-modal Contrast, we select 2-level matching in this paper to report the main results.

\begin{table}[t]

  \centering
  \caption{The investigation  of Hierarchical Cross-modal Contrastive Matching in Text-to-Video Retrieval.}
  \begin{tabular}{c|cccc}
  \toprule[1.5 pt] 
  
   {\textbf{Methods}}&{R@1} & {R@5} & {R@10}  &{MedR} \\ 
  
  \hline
  
  {\textbf{HiT}-\textit{sl}} &  23.5 & 56.2 & 68.8 & 4.0 \\
  
  {\textbf{HiT}-\textit{fl}} &  25.1 & 53.6 & 67.2 & 4.0 \\
  
   {\textbf{HiT}-\textit{4-level}} & 27.1 & \textbf{59.2} & 71.0 & \textbf{3.0}\\ 
  
  {\textbf{HiT}-\textit{3-level-a}} &   \textbf{28.5} &  58.4 & 71.0 & \textbf{3.0} \\ 
  
  {\textbf{HiT}-\textit{3-level-b}} &  26.7 & 58.5 & \textbf{71.4} & \textbf{3.0} \\ 
  
  {\textbf{HiT}} &  27.0 & 58.0 & 70.8 & \textbf{3.0} \\ 
  
  \toprule[1.5 pt]
   
\end{tabular}
 \label{abation1}
 \vspace{-3mm}
 \end{table}

\textbf{Momentum Cross-modal Contrast.}
To explore the impacts of the memory bank size, sufficient experiments are conducted. The results are shown in Table~\ref{tab5}. We vary the queue size of $K_v$ and $K_t$ from 0 to 8,192, and evaluate R@K and rsum. As shown in the results, it deserves attention that the introduction of large-scale negatives for similarity learning indeed achieves considerable performance improvements, in which we attribute it to broader negative sample interactions for obtaining more precise and discriminative representations. In addition, with the growth of queue size $K_v$ and $K_t$, retrieval performance is slightly degraded after the growth which is probably due to some positive samples are misclassified as negative samples.

\textbf{Momentum Encoders.}
For maintaining representation consistency in memory banks, we introduce two key encoders with momentum update for two modalities to generate representations. In this section, we abate two momentum encoders to explore their effectiveness in terms of maintaining representation consistency by evaluating the retrieval performance. We achieve the ablation by directly using query encoders to produce representations for memory banks. Table~\ref{momentum_encoder} presents the ablation results. We can find that it shows the performance degradation when we do not use momentum encoders. Particularly, it degrades performance at R@5 to 48.4\%, which clearly demonstrates the necessity of momentum encoders.

\begin{table}[h]

  \centering

  \caption{The impacts of Momentum Encoders for generating key representations.}
  
  \begin{tabular}{c|cccc}
  
   {\textbf{Encoders}}&{R@1} & {R@5} & {R@10}  &{MedR} \\ 
  
  \toprule[1.5 pt] 
  
  {Query Encoders} &  21.1 & 48.4 & 60.9 & 6.0\\ 
  
  {Key Encoders} &   \textbf{27.7} & \textbf{59.2} & \textbf{72.0}  & \textbf{3.0} \\

   
\end{tabular}
 \label{momentum_encoder}
 \end{table}

\textbf{Contrastive Loss.}
In Equation~\ref{b} and \ref{d}, InfoNCE is adopted as the Contrastive loss to perform common space learning. In this section, we use another commonly used loss function, \textit{i.e.,} Triplet Ranking Loss, as the objectives and present the retrieval performance for MSR-VTT in Table~\ref{tripletloss}. Though it exists the difficulty in tuning the appropriate combination of temperature and batch size, we find that InfoNCE achieves better performance than Triplet Ranking Loss in HiT.

\begin{table}[h]

  \centering
  \caption{The selection of Contrastive losses.}
  \begin{tabular}{c|cccc}
  
   {\textbf{Encoders}}&{R@1} & {R@5} & {R@10}  &{MedR} \\ 
  
  \toprule[1.5 pt] 
    {Triplet Ranking Loss}  & 25.6 & 56.7 & 69.1 & 4.0   \\
    
   {InfoNCE} &  \textbf{27.7} & \textbf{59.2} & \textbf{72.0}  & \textbf{3.0} \\

   
\end{tabular}
 \label{tripletloss}
 \vspace{-3mm}
 \end{table}
 
The temperature $\gamma$ in InfoNCE is a sensitive parameter. To show how $\gamma$ affects retrieval performance, the impacts of $\gamma$ with regard to rsum are presented in Table~\ref{tem}. We can observe that the best performance can be achieved when we set $\gamma$ to 0.07. A number with the same magnitude as 0.07 won't change the performance obviously.

\begin{table}[h]

\setlength{\belowcaptionskip}{-0.05cm} 
  \centering

  \caption{Parameter analysis for temperature $\gamma$.}

  \begin{tabular}{c|ccccc}

  {$\gamma$}&{0.0007} & {0.007} & { 0.07} & {0.7} & {7} \\

 \toprule[1 pt]
  

  {rsum} & 285.1  &311.2 & \textbf{320.3}  & 155.4 & 112.2 \\

 \end{tabular}
  \vspace{0mm}
 \label{tem}
 
 \end{table}

\textbf{Expert Utilization.}
In MSR-VTT, we use three types of expert embeddings as the visual input, including motion, appearance and audio features. The ablation of the different experts are in Table~\ref{feature}. 

\begin{table}[h]

  \centering

  \caption{Ablation study on different experts.}
  \begin{tabular}{c|cccc}

   {\textbf{Experts}}&{R@1} & {R@5} & {R@10}  &{MedR} \\ 
  
  \toprule[1.5 pt] 
  {Motion only} &  25.1 & 51.6 & 65.0 & 5.0 \\
 {Appearance only} &  18.2 & 41.9 & 55.5  & 6.0 \\
    {Audio only} &   10.9 & 22.1 & 31.1  & 16.0 \\
   {Motion + Appearance} &  24.2 & 52.5 & 65.1 & 5.0 \\
     {Motion + Audio} &  \textbf{28.1} & 57.8 & 71.5 & \textbf{3.0} \\
     {Appearance + Audio} &  20.1 &   46.9 & 58.7 & 5.0 \\
      {All} &  27.7 &  \textbf{59.2} &  \textbf{72.0} &  \textbf{3.0} \\
   
\end{tabular}
 \label{feature}
 \vspace{-4mm}
 \end{table}

From the results, we find that the motion expert achieves the best results when we only use one of three experts. Using audio features solely shows the worst performance. When using two experts, the combination of motion and audio experts achieves best results.  As analysed in \cite{mmt}, we also note that audio features contribute the most when being used together with others, which indicates that they provide many complementary cues.

\textbf{Feature Aggregation.}
As illustrated in Section~\ref{video_encoders} and ~\ref{text_encoders}, we leverage \textit{Average Pooling} to produce aggregated features before projection heads, in the sense of capturing important features from all tokens. Alternately, we evaluate three more aggregation methods, including \textit{Max Pooling}, \textit{1D-CNN} \cite{1dcnn} (kernel sizes: [2,3,4,5]) and using a \texttt{[CLS]} aggregated token. To obtain aggregated visual features from \texttt{[CLS]} token, similar to the text inputs, here we need to embed \texttt{[CLS]} and \texttt{[END]} tokens into the first and last positions of the visual input. We initialize them with random vectors. Table~\ref{Feature_Aggregation} presents comparison results in terms of text-video retrieval. Note that the decent results are not presented in \texttt{[CLS]}.  We suppose the reason is that the features are not well aggregated in the \texttt{[CLS]} at feature-level.
\vspace{-2mm}
\begin{table}[h]

  \centering
  \caption{Feature aggregation method comparison.}
  \begin{tabular}{c|cccc}
  
   {\textbf{Aggregation}}&{R@1} & {R@5} & {R@10}  &{MedR} \\ 
  
  \toprule[1.5 pt] 
  {\textit{Average Pooling}} & \textbf{27.7} & \textbf{59.2} & \textbf{72.0} & \textbf{3.0}  \\
 {\textit{Max Pooling}} &  26.8 & 60.1 & 71.2 & \textbf{3.0} \\
    {\textit{1D-CNN}} &  24.4 & 55.6 & 68.2 & 4.0  \\
   {\texttt{[CLS]}} &  24.2 & 53.1 & 65.0  & 5.0 \\

   
\end{tabular}
 \label{Feature_Aggregation}
 \vspace{-2mm}
 \end{table}


 \vspace{-4mm}
\section{Conclusion}
We summarize our paper in two aspects: 1) In Hierarchical Cross-modal Contrastive Matching, we show that taking advantage of feature hierarchies in transformers can achieve decent performance gains. 2) Momentum Cross-modal Contrast demonstrates that cross-modal learning can benefit from large-scale negative sample learning. For future: work: 1) To facilitate the exploitation of feature hierarchies in transformers, we can design the fusion modules to utilize hierarchical features more effectively and efficiently. 2)
To improve Momentum Cross-modal Contrast, some feature-level operations can be applied in memory banks, such as data mixing, hard negative selection, etc.
{\small
\bibliographystyle{ieee_fullname}
\bibliography{egbib}

\begin{thebibliography}{10}\itemsep=-1pt

\bibitem{Noise}
Elad Amrani, Rami Ben{-}Ari, Daniel Rotman, and Alex Bronstein.
\newblock Noise estimation using density estimation for self-supervised
  multimodal learning.
\newblock {\em CoRR}, abs/2003.03186, 2020.

\bibitem{AM}
Jingjing Chen, Chong{-}Wah Ngo, Fuli Feng, and Tat{-}Seng Chua.
\newblock Deep understanding of cooking procedure for cross-modal recipe
  retrieval.
\newblock In Susanne Boll, Kyoung~Mu Lee, Jiebo Luo, Wenwu Zhu, Hyeran Byun,
  Chang~Wen Chen, Rainer Lienhart, and Tao Mei, editors, {\em 2018 {ACM}
  Multimedia Conference on Multimedia Conference, {MM} 2018, Seoul, Republic of
  Korea, October 22-26, 2018}, pages 1020--1028. {ACM}, 2018.

\bibitem{hgr}
Shizhe Chen, Yida Zhao, Qin Jin, and Qi Wu.
\newblock Fine-grained video-text retrieval with hierarchical graph reasoning.
\newblock In {\em Proceedings of the IEEE/CVF Conference on Computer Vision and
  Pattern Recognition}, pages 10638--10647, 2020.

\bibitem{simclr}
Ting Chen, Simon Kornblith, Mohammad Norouzi, and Geoffrey Hinton.
\newblock A simple framework for contrastive learning of visual
  representations.
\newblock {\em arXiv preprint arXiv:2002.05709}, 2020.

\bibitem{improved}
Xinlei Chen, Haoqi Fan, Ross Girshick, and Kaiming He.
\newblock Improved baselines with momentum contrastive learning.
\newblock {\em arXiv preprint arXiv:2003.04297}, 2020.

\bibitem{simsiam}
Xinlei Chen and Kaiming He.
\newblock Exploring simple siamese representation learning.
\newblock {\em CoRR}, abs/2011.10566, 2020.

\bibitem{bertbase}
Jacob Devlin, Ming-Wei Chang, Kenton Lee, and Kristina Toutanova.
\newblock Bert: Pre-training of deep bidirectional transformers for language
  understanding.
\newblock {\em arXiv preprint arXiv:1810.04805}, 2018.

\bibitem{dual}
Jianfeng Dong, Xirong Li, Chaoxi Xu, Shouling Ji, Yuan He, Gang Yang, and Xun
  Wang.
\newblock Dual encoding for zero-example video retrieval.
\newblock In {\em Proceedings of the IEEE Conference on Computer Vision and
  Pattern Recognition}, pages 9346--9355, 2019.

\bibitem{corrae}
Fangxiang Feng, Xiaojie Wang, and Ruifan Li.
\newblock Cross-modal retrieval with correspondence autoencoder.
\newblock In {\em Proceedings of the 22nd ACM international conference on
  Multimedia}, pages 7--16, 2014.

\bibitem{ijcai20}
Zerun Feng, Zhimin Zeng, Caili Guo, and Zheng Li.
\newblock Exploiting visual semantic reasoning for video-text retrieval.
\newblock In Christian Bessiere, editor, {\em Proceedings of the Twenty-Ninth
  International Joint Conference on Artificial Intelligence, {IJCAI} 2020},
  pages 1005--1011. ijcai.org, 2020.

\bibitem{devise}
Andrea Frome, Greg~S Corrado, Jon Shlens, Samy Bengio, Jeff Dean, Marc'Aurelio
  Ranzato, and Tomas Mikolov.
\newblock Devise: A deep visual-semantic embedding model.
\newblock In {\em Advances in neural information processing systems}, pages
  2121--2129, 2013.

\bibitem{mmt}
Valentin Gabeur, Chen Sun, Karteek Alahari, and Cordelia Schmid.
\newblock Multi-modal transformer for video retrieval.
\newblock In {\em European Conference on Computer Vision (ECCV)}, 2020.

\bibitem{coot}
Simon Ging, Mohammadreza Zolfaghari, Hamed Pirsiavash, and Thomas Brox.
\newblock {COOT:} cooperative hierarchical transformer for video-text
  representation learning.
\newblock In {\em Advances in neural information processing systems}, 2020.

\bibitem{BYOL}
Jean{-}Bastien Grill, Florian Strub, Florent Altch{\'{e}}, Corentin Tallec,
  Pierre~H. Richemond, Elena Buchatskaya, Carl Doersch, Bernardo~{\'{A}}vila
  Pires, Zhaohan Guo, Mohammad~Gheshlaghi Azar, Bilal Piot, Koray Kavukcuoglu,
  R{\'{e}}mi Munos, and Michal Valko.
\newblock Bootstrap your own latent - {A} new approach to self-supervised
  learning.
\newblock In Hugo Larochelle, Marc'Aurelio Ranzato, Raia Hadsell,
  Maria{-}Florina Balcan, and Hsuan{-}Tien Lin, editors, {\em Advances in
  Neural Information Processing Systems 33: Annual Conference on Neural
  Information Processing Systems 2020, NeurIPS 2020, December 6-12, 2020,
  virtual}, 2020.

\bibitem{visualizingbert}
Yaru Hao, Li Dong, Furu Wei, and Ke Xu.
\newblock Visualizing and understanding the effectiveness of bert.
\newblock {\em arXiv preprint arXiv:1908.05620}, 2019.

\bibitem{he2020momentum}
Kaiming He, Haoqi Fan, Yuxin Wu, Saining Xie, and Ross Girshick.
\newblock Momentum contrast for unsupervised visual representation learning.
\newblock In {\em Proceedings of the IEEE/CVF Conference on Computer Vision and
  Pattern Recognition}, pages 9729--9738, 2020.

\bibitem{audio}
Shawn Hershey, Sourish Chaudhuri, Daniel P.~W. Ellis, Jort~F. Gemmeke, Aren
  Jansen, R.~Channing Moore, Manoj Plakal, Devin Platt, Rif~A. Saurous, Bryan
  Seybold, Malcolm Slaney, Ron~J. Weiss, and Kevin~W. Wilson.
\newblock {CNN} architectures for large-scale audio classification.
\newblock In {\em 2017 {IEEE} International Conference on Acoustics, Speech and
  Signal Processing, {ICASSP} 2017, New Orleans, LA, USA, March 5-9, 2017},
  pages 131--135. {IEEE}, 2017.

\bibitem{senet}
Jie Hu, Li Shen, Samuel Albanie, Gang Sun, and Enhua Wu.
\newblock Squeeze-and-excitation networks.
\newblock {\em {IEEE} Trans. Pattern Anal. Mach. Intell.}, 42(8):2011--2023,
  2020.

\bibitem{pixel}
Zhicheng Huang, Zhaoyang Zeng, Bei Liu, Dongmei Fu, and Jianlong Fu.
\newblock Pixel-bert: Aligning image pixels with text by deep multi-modal
  transformers.
\newblock {\em arXiv preprint arXiv:2004.00849}, 2020.

\bibitem{mochi}
Yannis Kalantidis, Mert~B{\"{u}}lent Sariyildiz, No{\'{e}} Pion, Philippe
  Weinzaepfel, and Diane Larlus.
\newblock Hard negative mixing for contrastive learning.
\newblock In Hugo Larochelle, Marc'Aurelio Ranzato, Raia Hadsell,
  Maria{-}Florina Balcan, and Hsuan{-}Tien Lin, editors, {\em Advances in
  Neural Information Processing Systems 33: Annual Conference on Neural
  Information Processing Systems 2020, NeurIPS 2020, December 6-12, 2020,
  virtual}, 2020.

\bibitem{SupCon}
Prannay Khosla, Piotr Teterwak, Chen Wang, Aaron Sarna, Yonglong Tian, Phillip
  Isola, Aaron Maschinot, Ce Liu, and Dilip Krishnan.
\newblock Supervised contrastive learning.
\newblock {\em CoRR}, abs/2004.11362, 2020.

\bibitem{1dcnn}
Yoon Kim.
\newblock Convolutional neural networks for sentence classification.
\newblock In Alessandro Moschitti, Bo Pang, and Walter Daelemans, editors, {\em
  Proceedings of the 2014 Conference on Empirical Methods in Natural Language
  Processing, {EMNLP} 2014, October 25-29, 2014, Doha, Qatar, {A} meeting of
  SIGDAT, a Special Interest Group of the {ACL}}, pages 1746--1751. {ACL},
  2014.

\bibitem{CCA}
Benjamin Klein, Guy Lev, Gil Sadeh, and Lior Wolf.
\newblock Associating neural word embeddings with deep image representations
  using fisher vectors.
\newblock In {\em {IEEE} Conference on Computer Vision and Pattern Recognition,
  {CVPR} 2015, Boston, MA, USA, June 7-12, 2015}, pages 4437--4446. {IEEE}
  Computer Society, 2015.

\bibitem{acitivity}
Ranjay Krishna, Kenji Hata, Frederic Ren, Li Fei{-}Fei, and Juan~Carlos
  Niebles.
\newblock Dense-captioning events in videos.
\newblock In {\em {IEEE} International Conference on Computer Vision, {ICCV}
  2017, Venice, Italy, October 22-29, 2017}, pages 706--715. {IEEE} Computer
  Society, 2017.

\bibitem{stacked}
Kuang-Huei Lee, Xi Chen, Gang Hua, Houdong Hu, and Xiaodong He.
\newblock Stacked cross attention for image-text matching.
\newblock In {\em Proceedings of the European Conference on Computer Vision
  (ECCV)}, pages 201--216, 2018.

\bibitem{CLIPBERT}
Jie Lei, Linjie Li, Luowei Zhou, Zhe Gan, Tamara~L. Berg, Mohit Bansal, and
  Jingjing Liu.
\newblock Less is more: Clipbert for video-and-language learning via sparse
  sampling.
\newblock In {\em {IEEE} Conference on Computer Vision}, 2021.

\bibitem{unicoder}
Gen Li, Nan Duan, Yuejian Fang, Ming Gong, Daxin Jiang, and Ming Zhou.
\newblock Unicoder-vl: A universal encoder for vision and language by
  cross-modal pre-training.
\newblock In {\em AAAI}, pages 11336--11344, 2020.

\bibitem{visualBERT}
Liunian~Harold Li, Mark Yatskar, Da Yin, Cho-Jui Hsieh, and Kai-Wei Chang.
\newblock Visualbert: A simple and performant baseline for vision and language.
\newblock {\em arXiv preprint arXiv:1908.03557}, 2019.

\bibitem{sigirLiuQGZY20}
Song Liu, Shengsheng Qian, Yang Guan, Jiawei Zhan, and Long Ying.
\newblock Joint-modal distribution-based similarity hashing for large-scale
  unsupervised deep cross-modal retrieval.
\newblock In {\em Proceedings of the 43rd International {ACM} {SIGIR}
  conference on research and development in Information Retrieval}, pages
  1379--1388, 2020.

\bibitem{CE}
Yang Liu, Samuel Albanie, Arsha Nagrani, and Andrew Zisserman.
\newblock Use what you have: Video retrieval using representations from
  collaborative experts.
\newblock In {\em 30th British Machine Vision Conference 2019, {BMVC} 2019,
  Cardiff, UK, September 9-12, 2019}, page 279. {BMVA} Press, 2019.

\bibitem{adamw}
Ilya Loshchilov and Frank Hutter.
\newblock Decoupled weight decay regularization.
\newblock In {\em 7th International Conference on Learning Representations,
  {ICLR} 2019, New Orleans, LA, USA, May 6-9, 2019}. OpenReview.net, 2019.

\bibitem{ViLBERT}
Jiasen Lu, Dhruv Batra, Devi Parikh, and Stefan Lee.
\newblock Vilbert: Pretraining task-agnostic visiolinguistic representations
  for vision-and-language tasks.
\newblock In {\em Advances in Neural Information Processing Systems}, pages
  13--23, 2019.

\bibitem{UniViLM}
Huaishao Luo, Lei Ji, Botian Shi, Haoyang Huang, Nan Duan, Tianrui Li, Xilin
  Chen, and Ming Zhou.
\newblock Univilm: {A} unified video and language pre-training model for
  multimodal understanding and generation.
\newblock {\em CoRR}, abs/2002.06353, 2020.

\bibitem{MEE}
Antoine Miech, Ivan Laptev, and Josef Sivic.
\newblock Learning a text-video embedding from incomplete and heterogeneous
  data.
\newblock {\em CoRR}, abs/1804.02516, 2018.

\bibitem{howto100m}
Antoine Miech, Dimitri Zhukov, Jean-Baptiste Alayrac, Makarand Tapaswi, Ivan
  Laptev, and Josef Sivic.
\newblock Howto100m: Learning a text-video embedding by watching hundred
  million narrated video clips.
\newblock In {\em Proceedings of the IEEE international conference on computer
  vision}, pages 2630--2640, 2019.

\bibitem{PIRL}
Ishan Misra and Laurens van~der Maaten.
\newblock Self-supervised learning of pretext-invariant representations.
\newblock In {\em 2020 {IEEE/CVF} Conference on Computer Vision and Pattern
  Recognition, {CVPR} 2020, Seattle, WA, USA, June 13-19, 2020}, pages
  6706--6716. {IEEE}, 2020.

\bibitem{LJE}
Niluthpol~Chowdhury Mithun, Juncheng Li, Florian Metze, and Amit~K.
  Roy{-}Chowdhury.
\newblock Learning joint embedding with multimodal cues for cross-modal
  video-text retrieval.
\newblock In Kiyoharu Aizawa, Michael~S. Lew, and Shin'ichi Satoh, editors,
  {\em Proceedings of the 2018 {ACM} on International Conference on Multimedia
  Retrieval, {ICMR} 2018, Yokohama, Japan, June 11-14, 2018}, pages 19--27.
  {ACM}, 2018.

\bibitem{infonce}
Aaron van~den Oord, Yazhe Li, and Oriol Vinyals.
\newblock Representation learning with contrastive predictive coding.
\newblock {\em arXiv preprint arXiv:1807.03748}, 2018.

\bibitem{iclr21}
Mandela Patrick, Po{-}Yao Huang, Yuki~Markus Asano, Florian Metze, Alexander~G.
  Hauptmann, Jo{\~{a}}o~F. Henriques, and Andrea Vedaldi.
\newblock Support-set bottlenecks for video-text representation learning.
\newblock In {\em ICLR}, 2021.

\bibitem{view2}
Matthew Peters, Mark Neumann, Luke Zettlemoyer, and Wen-tau Yih.
\newblock Dissecting contextual word embeddings: Architecture and
  representation.
\newblock In {\em EMNLP}, 2018.

\bibitem{confsigirQianWHFX21}
Shengsheng Qian, Jinguang Wang, Jun Hu, Quan Fang, and Changsheng Xu.
\newblock Hierarchical multi-modal contextual attention network for fake news
  detection.
\newblock In {\em the 44th International {ACM} {SIGIR} Conference on Research
  and Development in Information Retrieval}, pages 153--162, 2021.

\bibitem{aaaiQianXZFX21}
Shengsheng Qian, Dizhan Xue, Huaiwen Zhang, Quan Fang, and Changsheng Xu.
\newblock Dual adversarial graph neural networks for multi-label cross-modal
  retrieval.
\newblock In {\em Thirty-Fifth {AAAI} Conference on Artificial Intelligence,
  {AAAI}}, pages 2440--2448, 2021.

\bibitem{understanding}
Yifan Qiao, Chenyan Xiong, Zhenghao Liu, and Zhiyuan Liu.
\newblock Understanding the behaviors of bert in ranking.
\newblock {\em arXiv preprint arXiv:1904.07531}, 2019.

\bibitem{LSMDC}
Anna Rohrbach, Marcus Rohrbach, Niket Tandon, and Bernt Schiele.
\newblock A dataset for movie description.
\newblock In {\em Proceedings of the IEEE conference on computer vision and
  pattern recognition}, pages 3202--3212, 2015.

\bibitem{AVLnet}
Andrew Rouditchenko, Angie~W. Boggust, David Harwath, Dhiraj Joshi, Samuel
  Thomas, Kartik Audhkhasi, Rog{\'{e}}rio Feris, Brian Kingsbury, Michael
  Picheny, Antonio Torralba, and James~R. Glass.
\newblock Avlnet: Learning audio-visual language representations from
  instructional videos.
\newblock {\em CoRR}, abs/2006.09199, 2020.

\bibitem{pvse}
Yale Song and Mohammad Soleymani.
\newblock Polysemous visual-semantic embedding for cross-modal retrieval.
\newblock In {\em Proceedings of the IEEE Conference on Computer Vision and
  Pattern Recognition}, pages 1979--1988, 2019.

\bibitem{vlBERT}
Weijie Su, Xizhou Zhu, Yue Cao, Bin Li, Lewei Lu, Furu Wei, and Jifeng Dai.
\newblock Vl-bert: Pre-training of generic visual-linguistic representations.
\newblock {\em arXiv preprint arXiv:1908.08530}, 2019.

\bibitem{sun2019learning}
Chen Sun, Fabien Baradel, Kevin Murphy, and Cordelia Schmid.
\newblock Learning video representations using contrastive bidirectional
  transformer.
\newblock {\em arXiv preprint arXiv:1906.05743}, 2019.

\bibitem{videoBERT}
Chen Sun, Austin Myers, Carl Vondrick, Kevin Murphy, and Cordelia Schmid.
\newblock Videobert: A joint model for video and language representation
  learning.
\newblock In {\em Proceedings of the IEEE International Conference on Computer
  Vision}, pages 7464--7473, 2019.

\bibitem{lxmert}
Hao Tan and Mohit Bansal.
\newblock Lxmert: Learning cross-modality encoder representations from
  transformers.
\newblock {\em arXiv preprint arXiv:1908.07490}, 2019.

\bibitem{rediscovers}
Ian Tenney, Dipanjan Das, and Ellie Pavlick.
\newblock Bert rediscovers the classical nlp pipeline.
\newblock {\em arXiv preprint arXiv:1905.05950}, 2019.

\bibitem{view1}
Ian Tenney, Dipanjan Das, and Ellie Pavlick.
\newblock Bert rediscovers the classical nlp pipeline.
\newblock In {\em ACL}, 2019.

\bibitem{CMC}
Yonglong Tian, Dilip Krishnan, and Phillip Isola.
\newblock Contrastive multiview coding.
\newblock {\em CoRR}, abs/1906.05849, 2019.

\bibitem{infomin}
Yonglong Tian, Chen Sun, Ben Poole, Dilip Krishnan, Cordelia Schmid, and
  Phillip Isola.
\newblock What makes for good views for contrastive learning.
\newblock {\em CoRR}, abs/2005.10243, 2020.

\bibitem{BERT}
Ashish Vaswani, Noam Shazeer, Niki Parmar, Jakob Uszkoreit, Llion Jones,
  Aidan~N Gomez, {\L}ukasz Kaiser, and Illia Polosukhin.
\newblock Attention is all you need.
\newblock In {\em Advances in neural information processing systems}, pages
  5998--6008, 2017.

\bibitem{view3}
Jesse Vig.
\newblock A multiscale visualization of attention in the transformer model.
\newblock In {\em ACL}, 2019.

\bibitem{adversarial}
Bokun Wang, Yang Yang, Xing Xu, Alan Hanjalic, and Heng~Tao Shen.
\newblock Adversarial cross-modal retrieval.
\newblock In {\em Proceedings of the 25th ACM international conference on
  Multimedia}, pages 154--162, 2017.

\bibitem{tmm}
Wei Wang, Junyu Gao, Xiaoshan Yang, and Changsheng Xu.
\newblock Learning coarse-to-fine graph neural networks for video-text
  retrieval.
\newblock {\em IEEE Transactions on Multimedia}, PP:1--1, 07 2020.

\bibitem{position}
Yaxiong Wang, Hao Yang, Xueming Qian, Lin Ma, Jing Lu, Biao Li, and Xin Fan.
\newblock Position focused attention network for image-text matching.
\newblock {\em arXiv preprint arXiv:1907.09748}, 2019.

\bibitem{frag}
Yiling Wu, Shuhui Wang, Guoli Song, and Qingming Huang.
\newblock Learning fragment self-attention embeddings for image-text matching.
\newblock In {\em Proceedings of the 27th ACM International Conference on
  Multimedia}, pages 2088--2096, 2019.

\bibitem{unsupervised}
Zhirong Wu, Yuanjun Xiong, X~Yu Stella, and Dahua Lin.
\newblock Unsupervised feature learning via non-parametric instance
  discrimination.
\newblock In {\em Proceedings of the IEEE Conference on Computer Vision and
  Pattern Recognition}, 2018.

\bibitem{momerybank}
Zhirong Wu, Yuanjun Xiong, Stella Yu, and Dahua Lin.
\newblock Unsupervised feature learning via non-parametric instance-level
  discrimination.
\newblock In {\em 2018 IEEE/CVF Conference on Computer Vision and Pattern
  Recognition}, 2018.

\bibitem{T2VLAD}
Yi~Yang Xiaohan~Wang, Linchao~Zhu.
\newblock T2vlad:global-local sequence alignment for text-video retrieval.
\newblock In {\em 2021 {IEEE} Conference on Computer Vision}, 2021.

\bibitem{s3d}
Saining Xie, Chen Sun, Jonathan Huang, Zhuowen Tu, and Kevin Murphy.
\newblock Rethinking spatiotemporal feature learning: Speed-accuracy trade-offs
  in video classification.
\newblock In Vittorio Ferrari, Martial Hebert, Cristian Sminchisescu, and Yair
  Weiss, editors, {\em Computer Vision - {ECCV} 2018 - 15th European
  Conference, Munich, Germany, September 8-14, 2018, Proceedings, Part {XV}},
  volume 11219 of {\em Lecture Notes in Computer Science}, pages 318--335.
  Springer, 2018.

\bibitem{graph}
Yu Xiong, Qingqiu Huang, Lingfeng Guo, Hang Zhou, Bolei Zhou, and Dahua Lin.
\newblock A graph-based framework to bridge movies and synopses.
\newblock In {\em Proceedings of the IEEE International Conference on Computer
  Vision}, pages 4592--4601, 2019.

\bibitem{msrvtt}
Jun Xu, Tao Mei, Ting Yao, and Yong Rui.
\newblock Msr-vtt: A large video description dataset for bridging video and
  language.
\newblock In {\em Proceedings of the IEEE conference on computer vision and
  pattern recognition}, pages 5288--5296, 2016.

\bibitem{jsfusion}
Youngjae Yu, Jongseok Kim, and Gunhee Kim.
\newblock A joint sequence fusion model for video question answering and
  retrieval.
\newblock In {\em Proceedings of the European Conference on Computer Vision
  (ECCV)}, pages 471--487, 2018.

\bibitem{CT-SAN}
Youngjae Yu, Hyungjin Ko, Jongwook Choi, and Gunhee Kim.
\newblock End-to-end concept word detection for video captioning, retrieval,
  and question answering.
\newblock In {\em 2017 {IEEE} Conference on Computer Vision and Pattern
  Recognition, {CVPR} 2017, Honolulu, HI, USA, July 21-26, 2017}, pages
  3261--3269. {IEEE} Computer Society, 2017.

\bibitem{HSE}
Bowen Zhang, Hexiang Hu, and Fei Sha.
\newblock Cross-modal and hierarchical modeling of video and text.
\newblock In Vittorio Ferrari, Martial Hebert, Cristian Sminchisescu, and Yair
  Weiss, editors, {\em Computer Vision - {ECCV} 2018 - 15th European
  Conference, Munich, Germany, September 8-14, 2018, Proceedings, Part {XIII}},
  volume 11217 of {\em Lecture Notes in Computer Science}, pages 385--401.
  Springer, 2018.

\bibitem{actBERT}
Linchao Zhu and Yi Yang.
\newblock Actbert: Learning global-local video-text representations.
\newblock In {\em Proceedings of the IEEE/CVF Conference on Computer Vision and
  Pattern Recognition}, pages 8746--8755, 2020.

\end{thebibliography}
}

\end{document}